\documentclass{svproc}
\usepackage{url}

\usepackage{graphicx}
\usepackage{color}
\usepackage{amsmath}
\usepackage{amsfonts}
\usepackage{mathrsfs}
\usepackage{sidecap}
\usepackage{subcaption} 
\usepackage{comment}
\usepackage[utf8]{inputenc}
\usepackage[english]{babel}
\usepackage{tabularx}

\usepackage{flushend}

\usepackage{graphics}
\usepackage{caption}
\usepackage{hyperref}
\usepackage{subfiles}
\usepackage{blindtext}
\usepackage{subfiles}
\usepackage{blindtext}

\begin{document}
\mainmatter              
\title{Learning Parameters for Balanced Index Influence Maximization}
\titlerunning{Learning Parameters for IM}  
%
\author{Manqing Ma\inst{1} \and Gyorgy Korniss\inst{1} \and  Boleslaw K. Szymanski\inst{1}}
\authorrunning{M. Ma, G. Korniss, and B.K. Szymanski} 
%
\tocauthor{}
\institute{$^1$Network Science and Technology Center, Rensselaer Polytechnic Institute, Troy, NY\\
\email{\{mam6, szymab, korniss\}@rpi.edu}\\ 
}

\maketitle              

\begin{abstract}
Influence maximization is the task of finding the smallest set of nodes to be activated in a social network such that their aggregated influence can trigger an activation cascade that reaches the targeted network coverage, where threshold rules determine the outcome of influence. This problem is NP-hard and it has generated a significant amount of recent research on finding efficient heuristics. 
We focus on a {\it Balance Index} algorithm that relies on three parameters to tune its performance to the given network structure. We propose using a supervised machine-learning approach for such tuning. We select the most influential graph features for the parameter tuning. Then, using random-walk-based graph sampling, we create small snapshots from the given synthetic and large-scale real-world networks. Using exhaustive search, we find for these snapshots the high accuracy values of BI parameters to use as a ground truth. Then, we train our machine-learning model on the snapshots and apply this model to the real-word network to find the best BI parameters. We apply these parameters to the sampled real-world network to measure the quality of the initiator sets found this way. We use various real-world networks to successfully validate our approach against other heuristic.
.
\end{abstract}
%
\section{Introduction}
In a social network setting, influence maximization (IM) is a task motivated by viral marketing. Its goal is to identify the smallest set of social network nodes, which if initially activated to a new state, will collectively influence others to activate. Originally defined by Kempe \textit{et. al.} \cite{kempe2003maximizing}, the problem assumes the known directed social network with either weighted or unweighted edges, a stochastic influence propagation model (i.e., the Linear Threshold Model (LTM)  \cite{chen2010scalable}, in which threshold rules determine influence outcome), and the challenge is to find the minimal {\it set initiators} that maximize the spread of their initiated state. The corresponding influence maximization problem is NP-hard \cite{kempe2003maximizing} and it has generated a significant amount  of recent research on finding efficient heuristics. One approach focuses on various node indexing heuristics \cite{kempe2005influential,kitsak2010identification}, in which all nodes in the graph are indexed based on their properties, and the highest ranking nodes are selected to the seed set. In this approach graph features related percolation, such as degree, or betweenness, are often used for indexing \cite{morone2015influence,pei2017efficient,Karsai2016local,Karsai2018Threshold}. 
        

From the application perspective, it helps to include the specific context information into the node indexing heuristic. In a survey paper \cite{li2018influence}, the authors concluded that the IM challenge includes finding how the graph structure affects the solution and how to identify a robust seed set given a limited number of graph changes. To address this challenge, we use a {\it Balance Index} (BI) algorithm \cite{karampourniotis2019influence} that relies on three parameters to tune its performance to the given network structure. Here, we propose to use Machine-Learning (ML) for such tuning. 

We use a standard supervised ML approach in which the ML model learns from the training data, validates the model performance on the test data, and predicts the best parameters for the given network. 
In summary, we use ML to find the most influential graph features and apply them to the parameter tuning for the BI algorithm. 

The main contributions of this work are as follows. We propose a random-walk-based graph sampling that quickly creates snapshots of large scale real-world networks as training data. We developed a method of finding the most influential graph features for the BI algorithm. We also validated the applicability of the synthetic network trained ML model to various real-world networks.

%

%

\section{Methodology}\label{method}
We use the following notation. We consider a social network with $N$ nodes and set of $E$ edges, so with $|E|$ edges, undergoing conversion from the old to new state using Threshold Model spread process. We denote by $r_i$ the resistance of node $i$ to spreading, which is the number of neighbors of $i$ that needs to turn active in order for node $i$ to become active. Each node in the network has a fractional threshold for activation, that represents the node’s resistance to peer pressure. The spreading rule is that an inactive node $i$, with in-degree $k^{in}_i$ and threshold $\phi_i$, is activated by its in-neighbors only when the their fraction of activated nodes is higher than the node’s threshold, that is $\sum_{j\in N_i} 1 \geq \phi_ik^{in}_i$, where $N_i$ denotes set of neighbors of node $i$. The is deterministic process and once a node is activated, it cannot return to its previous state. 
In addition, $k^{out}_i$ stands for out-degree of node $i$, which represents the immediate decrease of the network resistance to spread when $i$ is activated, and $k^{out,1}_i$ is the number of $i$ neighbors with resistance 1, which means that once $i$ is activated all of these neighbors will be immediately activated as well, so this value represent the immediate increase in the number of activated nodes when $i$ is activated.

The Balanced Index (BI) introduced in \cite{karampourniotis2019influence} quantifies the combined potential of being effective initial spreader based on node's resistance, out-degree, and the number of out-neighbors ready for activation with resistance 1, using parameters defined as:
\begin{equation}\label{eq:BI}
    BI_i = a r_i + b k^{out}_i + c \sum_{j \in \partial i | r_j = 1} (k^{out}_j - 1) 
\end{equation}
where $a + b + c = 1$ and $a, b, c \geq 0$.

Given a large social network, attempting to find effective parameters for applying the BI algorithm to this network would be prohibitively expensive. 
So, our approach first creates many of its subgraphs to avoid random variance in their quality and then uses a supervised classification task to find those parameters. Next, the averaged parameters are applied to the original graph. Here, we use a number of real-world networks, instead of just one, to measure efficiency of our approach for each of these graphs.

 \subsection{Random-walk graph sampling and supervised classification task}
    
We need a graph sampling method that could create subgraphs that are similar to the original graph in features relevant to the values of the BI parameters. Many graph sampling methods were tested for the similarity between the original network and the resulting subgraphs in \cite{Backstrom2011supervised}. The author found that the random-walk sampling preserves the structural graph features well. This conclusion motivates us to use the random-walk sampling in our approach.
    
Each sample is created in one complete walk with no restarts to ensure the created subgraph is fully connected.

We denote the dimension of input space (also known as feature space) of this task  as $n$. Here, $n$ is the number of graph features selected for our task and the feature space is $R^n$. Each feature vector $x_i$ is represented as $x_i = (x_i^{(1)}, x_i^{(2)}, ..., x_i^{(n)})$. The output (target) space of dimension $m$ is defined as $y_i = (y_i^{(1)}, y_i^{(2)}, ..., y_i^{(m)})$. The targets could be further sliced into classes $\{C_j\}_l$, enabling us to transform our task to a multi-class or binary (two-class) classification problem.

Given the dataset of size $D$, we split it into two disjoint parts. The training dataset of size $M$ is represented as $T = {(x_1, y_1), (x_2, y_2), ..., (x_{M}, y_{M})}$, and the complementary testing dataset is represented as $V = {(x_{M + 1}, y_{M + 1}), ..., (x_D, y_D)}$.

Several methods of classification have shown good performance for a small number of features, including the Logistic Regression Classification and Random Forest Classification\cite{malik2012}.
We chose the Random Forest Classification method for our task because of its high adaptability to input scales, input noise and fitting to both linear and nonlinear problems with no precedence hypotheses.

\subsection{Datasets and baseline comparison}

We use two types original networks on which we want to run BA algorithm, the synthetic ER graphs with edge swapping, and real-world networks. For both types, we generate sample subgraphs for model training. All those networks are summarized in Table \ref{tb:dataset}.
            \begin{table}[ht!]
            \small
            \caption{Listing of Datasets Used for Sample Generation for Learning and Testing}
           \begin{center}
          \begin{tabular}{|l|c|r|}
            \hline
            \multicolumn{3}{|c|}{Synthetic Networks}\\
            \hline
                Network Generation Model & Parameters & Count\\ 
            \hline
               ER with edge swapping     & N = 100, k = 5, 10          & 75 * 2                  \\
                                         & N = 300, k = 5, 10          & 75 * 2                  \\
                                         & N = 500, k = 5, 10          & 75 * 2                  \\
        \hline
            \multicolumn{3}{|c|}{Real-World Networks}\\
            \hline
                Network name & Parameters & Count\\
            \hline
                Amazon Co-purchasing network samples& N $\sim$ 1000 & 1000 \\
                Twitter retweet network samples: ``center'' & N $\sim$ 1000 & 50                     \\ 
                Twitter retweet network samples: ``lean left'' & N $\sim$ 1000    & 50                     \\ 
                Facebook network samples   & N $\sim$ 500 & 50\\ 
                CA-CondMat network samples & N $\sim$ 500 & 50\\ 
                CA-HepPh network samples   & N $\sim$ 500 & 50\\
            \hline
           \end{tabular}
           \end{center} 
           \label{tb:dataset}
            \end{table}
            

In \cite{Bounova2012overview}, the authors list 30 important graph metrics that collectively characterize the graph structure. Selecting the metrics that need to be preserved in the subgraphs created from the original graph, we need to take into account their influence on the efficient values of the BI parameters and the complexity of computing them, since they need to be computed on the original graph. 
In the BI algorithm, the parameters represent the importance of node's features that are related to each node degree, resilience, and the number of out-neighbors that are ready for immediate activation. Hence, these relations can be captured using other state-of-art node's metrics, such as the density of neighborhood, the mean and variance of in and out degrees, or the graph's ``degree assortativity''\cite{newman2003mixing}. In addition to the graph features intrinsic to the graph structure, the threshold $\phi$ distribution and the targeted cascade coverage (i.e., the targeted fraction of nodes to be influenced) also affect the efficient value of the BI parameters.

Taking into account the complexity and influence of the metrics on the efficient values of the BI parameters, we selected the features listed in Table \ref{tb:graph_features}.     
\begin{table}
    \small
    \caption{Graph features, where $C$ denotes a local clustering coefficient, $N^{out}_i$ stands for average out-degree of neighbors of node $i$, $cov$ is the targeted coverage of a cascade, and $\rho$ denotes an out-degree assortativity of the graph. The mean and standard deviation of a distribution of values $v$ are denoted as $\bar{v}, \sigma_v$, respectively.  }
    \begin{center}
    \begin{tabular}{|c|c|c|}
    \hline
    Feature         & Definition     & Complexity\\
    \hline
    $N$  & Node count & - \\
    \hline
    $\bar{C}$  & $\frac{1}{N} \sum_{i = 1}^{N} \frac{|\{e_{jk}: v_j, v_k \in \delta_i, e_{jk} \in E\}|}{k^{out}_i(k^{out}_i - 1)}$ 
                                     & $O(N\overline{k^{out}}^2)$\\
    \hline
    $\sigma_C$     & $\sqrt{\frac{1}{N} \sum_{i = 1}^{N} (C_i - \bar{C}) ^ 2}, C_i = \frac{|\{e_{jk}: v_j, v_k \in \delta_i, e_{jk} \in E\}|}{k^{out}_i(k^{out}_i - 1)}$                             & $O(N\overline{k^{out}} ^ 2)$\\
    \hline
    $\overline{k^{out}}$ 
                    & $\frac{1}{N} \sum_{i = 1}^{N} k^{out}_i$ 
                                     & $O(N)$\\
    \hline
    $\sigma_{k^{out}}$ 
                    & $\frac{1}{N} \sqrt{\sum_{i=1}^{N} (k^{out}_i-\overline{k^{out}})^2}$ 
                                     & $O(N)$\\
    \hline
    $\overline{N^{out}}$ 
                    &
                    $\frac{1}{N} \sum_{i = 1}^{N} \frac{1}{|\delta_{i}|}\sum_{j \in \delta_{i}}k^{out}_j$
                                     &$O(N\overline{k^{out}})$\\
    \hline
    $\sigma_{N^{out}}$ 
                    & $\frac{1}{N} \sum_{i = 1}^{N} \sqrt{ \frac{1}{|\delta_{i}|}\sum_{j \in \delta_{i}}
                    (k^{out}_j - \overline{N^{out}})^2}$
                    & $O(N\overline{k^{out}})$ \\
    \hline
    $\rho$          & See Eq[21] in \cite{newman2003mixing} 
                                     & See \cite{newman2003mixing}\\
    \hline
    $E_d$           & $\frac{|E|}{\binom{N}{2}}$
                                     & $O(N\overline{k^{out}})$\\
    \hline
    $cov$          & $\frac{N^t}{N}$ & $O(1)$ \\
    \hline
    $\bar{\phi}$ & $\frac{1}{N} \sum_{i = 1}^N \phi_i$ 
                                     & $O(N)$\\
    \hline
    $\sigma_{\phi}$    & $\frac{1}{N} \sum_{i = 1}^N (\phi_i - \bar{\phi})^2$  & $O(N)$ \\
    \hline
    \end{tabular} \label{tb:graph_features}
    \end{center}
\end{table}

        
By generating synthetic graphs based on random graphs, we get a dataset covering a broad range of graph features, so we expect that the BI parameters values obtained with them will perform worse on real-world networks than the parameters obtained by real-word network sampling. We expect that subgraph generated from a real-world network will preserve well its graph structure characteristics.

For finding the ground-truth best parameter values in each subgraph setting (cascade coverage and threshold distribution), we simply perform the search over the triangle grid of $\lceil{\frac{\max(a)}{2prec}+1\rceil} \times \lceil{\frac{\max(b)}{prec}+1\rceil}$ points, where $prec=0.01$, so this is a triangle grid of $51\times 101$ points, which require $5,151$ executions of indexing of the nodes with complexity in the order of $O(N\overline{k^{out}})$, and then running the spread that also requires $O(N\overline{k^{out}})$ steps. Then, the best values of $a$ and $b$ are selected for generating the smallest number of initiators.
            
The synthetic dataset is split using $M=2D/3$, so 2/3 of data for training and 1/3 for testing. After the model is trained on synthetic dataset, it is validated using the real-world network data.

The performance of each solution is measured using the number of initiators needed by this solution to reach the targeted network coverage, so smaller measurement indicates better performance.  
        
        We compared the tuned BI heuristic with the following node indexing based heuristics:

        \begin{enumerate}
            \item $res$: Node resistance based indexing, which corresponds for the BI with the values for (a;b;c) equal to (1;0;0).
            \item $deg$: Adaptive high out-degree based indexing \cite{kitsak2010identification} corresponding to the BI with the parameter values set to (0;1;0).
            \item $RD$: Resistance and node out-degree based heuristic strategy, corresponding to the BI with the parameter values set to (0.5;0.5;0).
            \item $CI-TM$: Collective influence based indexing for a sphere of influence when $L=1$ \cite{morone2015influence}. Since the metric of CI-TM is only composed of the out-degree of the nodes surrounding the target node, so this sets the BI parameter values to (0;0.5;0.5).  \cite{karampourniotis2019influence}.
            
        \end{enumerate}

\section{Result and Analysis}\label{result}
            Here, we first examine the relationship between different parameter values and cases in which they deliver their best performance. 

            % ...
            
\begin{figure}[ht!]
                \centering
                \includegraphics[width=1 \textwidth]{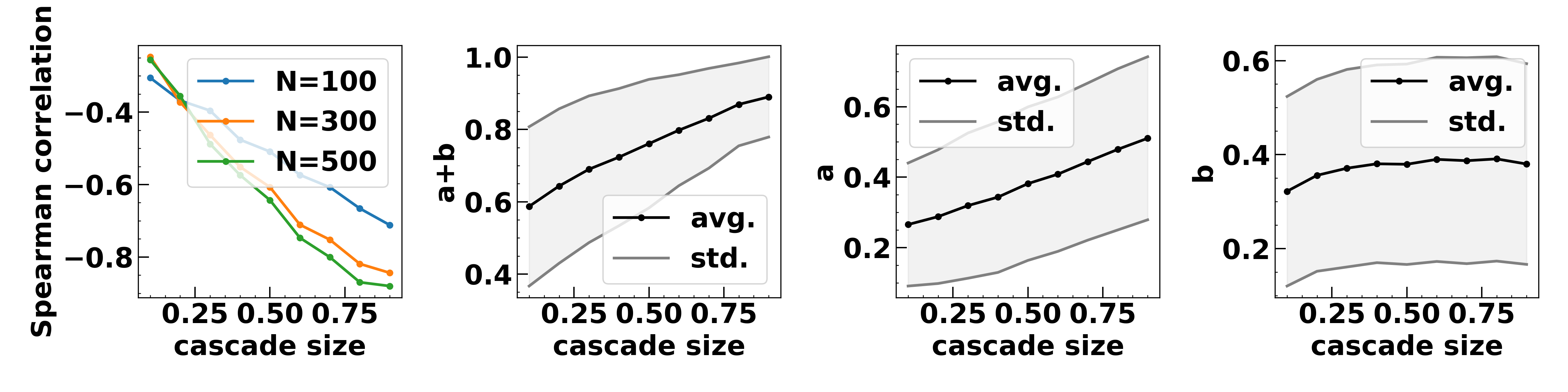}
                \caption{Optimal values obtained in a triangle grid search with precision $0.01$ over synthetic subgraphs and the different targeted cascade coverage values. (Left) The Spearman correlation between the best values of $a$ and $b$. (Left-Center) Sum of the best $a$ and $b$. (Right-Center) The best $a$. (Right) the best $b$}
                \label{ab_corr}
            \end{figure}

            For the synthetic subgraphs and given the range of targeted cascade coverage, Fig. \ref{ab_corr} shows the optimal $a$ and $b$ values in a triangle grid search with precision $0.01$ (so with $51\times 101=5,151$ points), while the third parameters is set as $c=1-a-b$. 
            
The first plot of Fig. \ref{ab_corr} shows that as the network size increases, the plot moves toward the diagonal. There is also an increase of the negative correlation between the best values of $a$ and $b$ when the targeted cascade coverage increases, together with the increase of their sum to one, shown in the second plot. 
The third and fourth plots show the increasing importance of out-degree($b$) and resistance($a$) when the larger cascade coverage is needed. 
The conclusion is that the larger is the targeted cascade coverage, the less important is to focus on ready for immediate activation out-neighbors and to concentrate instead on the long-term strategy of selecting the most resistant ($a$) and influential ($b$) nodes.  
           \begin{figure}
                \centering
                \includegraphics[width= \columnwidth]{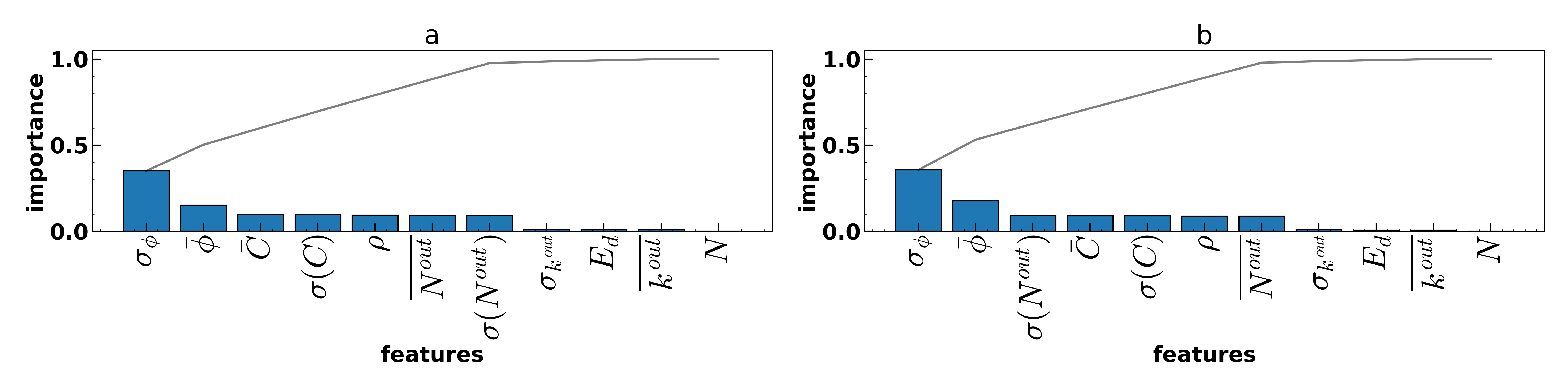}
                \caption{Feature importance (For $cov=0.9$). (Left) Bars from the left to the right show features in the order of importance for the BI coefficient $a$: $\sigma_{\phi}$, standard deviation; $\bar{\phi}$, average value of threshold; $\bar{C}$ and $\sigma(C)$, average and standard deviation of local clustering coefficient; $\rho$, assortativity; $\bar{N^{out}}$ and $\sigma_{N^{out}}$, average and standard deviation of average out-degree of neighbors; $\sigma_{k^{out}}$, standard deviation of out-degree; $E_d$, edge density; $\bar{k^{out}}$, average out-degree; and $N$, the number of nodes. (Right) For coefficient $b$, bars show the same features but in the order of significance for the BI $b$ coefficient. In both plots, the lines above the bars show cumulative importance of features below and to the left of a point of reference}
                \label{fig:feature_importance}
            \end{figure}
        \subsection{Identifying the most important features}
Although the classification model can be used as a black-box, knowing the important features may reduce or increase feature dimension. 
For the Random Forest model, the feature importance corresponds to the cumulative entropy reduction as each feature is a root of the sub-tree in the decision tree of the forest.
Fig. \ref{fig:feature_importance} shows the results of both classification tasks on synthetic subgraphs. The plots show that $\sigma_{\phi}$ is dominant among all features by capturing over $30\%$ of the overall importance. The second is $\bar{\phi}$ that claims over $15\%$ of importance. The next five features account each for nearly 10\% of importance, while the remaining four are negligible.  
 
\subsection{Training the classification tasks on synthetic subgraphs}
\begin{figure}[ht!]
                \centering
                \includegraphics[width= \textwidth]{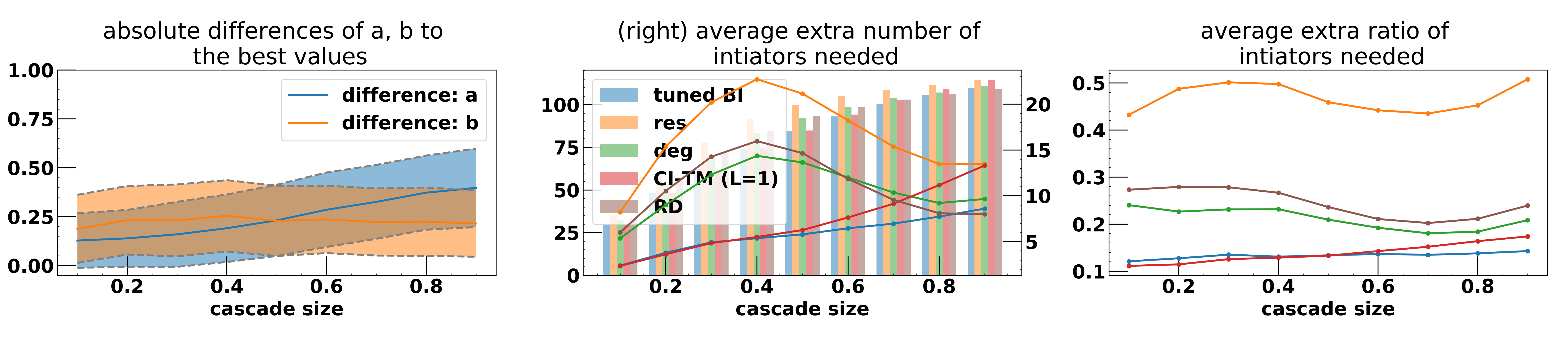}
                \caption{Comparison of performance of BI parameters found by model trained on synthetic subgraphs to the other node indexing based heuristics. (Left) Range of difference between $a$ and $b$ parameters found by the model and by exhaustive search. (Center) Bars show size initiation set for each heuristic with scale on the left. Plots show additional initiators needed by each heuristic over what was required by BI parameters found by exhaustive search with scale on the right. (Right) Fraction of the best initiation set needed by each heuristic to achieve the same coverage. Our tuned BI heuristic requires the smallest such fraction, with CI-TM matching it for smaller cascades.}
                \label{fig:ab_syn}
            \end{figure}

Fig. \ref{fig:ab_syn} shows the Random Forest Classification performance on the synthetic subgraphs. The first subplot shows the absolute differences between the predicted and optimal values of $a$ and $b$ obtained by a triangle grid search. The difference is less than $0.2$ on both sides of the optimal values. The padding shows boundaries of single standard deviation from the average line. The second and the third subplots compare performance of our method with other node indexing based heuristics. 
The bar plot in the second plots show the total number of initiators, while the line plots chart the numbers of initiators needed by heuristics over the optimal number of initiators. The third plot shows the fraction of additional initiators needed by heuristic compared to such fraction when $a$ and $b$ values obtained by the triangle grid search are used. 
                    \begin{figure}[ht!]
                \centering
                 \includegraphics[width = \textwidth]{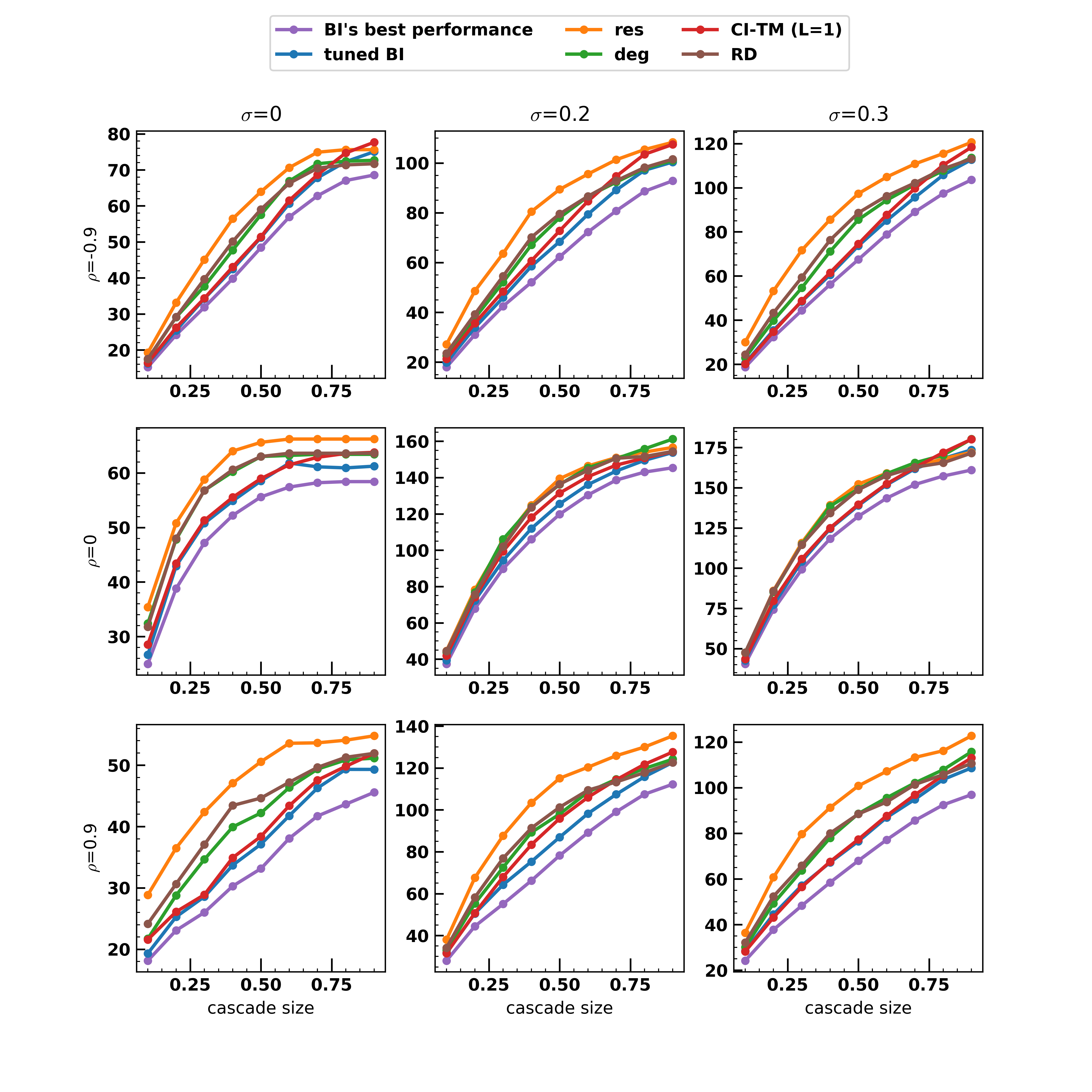}
                \caption{Number of initiators needed for a range of values for cascade coverage and different node ranking metrics on the synthetic subgraphs.  Each plot compares heuristics for different ranges of threshold's standard deviation $\sigma_{\phi}$ $\sigma$ in the plot) and assortativity $\rho$}
                \label{fig:er_9plots}
            \end{figure}    
The synthetic networks contain a spectrum of network features \ref{fig:er_9plots}. Hence, we want to see if the result averaged over different network realizations, characterized by varying degree assortativity, $\rho$, and threshold $\phi$ distribution standard deviation $\sigma_{\phi}$. Similarly, comparing the results at each targeted cascade coverage shown in Figs \ref{fig:er_9plots} and  \ref{fig:ab_syn} shows that our method ``tuned BI'' performs second only to the exhaustive triangle grids search labeled as ``best performance BI''. In most cases when cascade size is small ``CI-TM'' with $L=1$ is performs comparably with ``tuned BI''. The next two best performing approaches include ``deg'' and ``RD'', while ``res'' generally perform the worst.

In summary, the results show that parameter tuning using our Random Forest Classification has achieved a convincing performance boost on the synthetic dataset.
            
        \subsection{Validating the approach on real-world networks}

We used the model trained on synthetic networks for the real-world graphs (subgraph samples) that the graph metric values are not known beforehand. For the Amazon co-purchasing network subgraph samples, we further performed the grid search with the predicted $a$ and $b$ and compared to results with $a$ and $b$ found by the grid search. The difference was smaller than $0.05$ for both parameters.

\begin{figure}[ht!]
    \centering
    \includegraphics[width= \textwidth]{{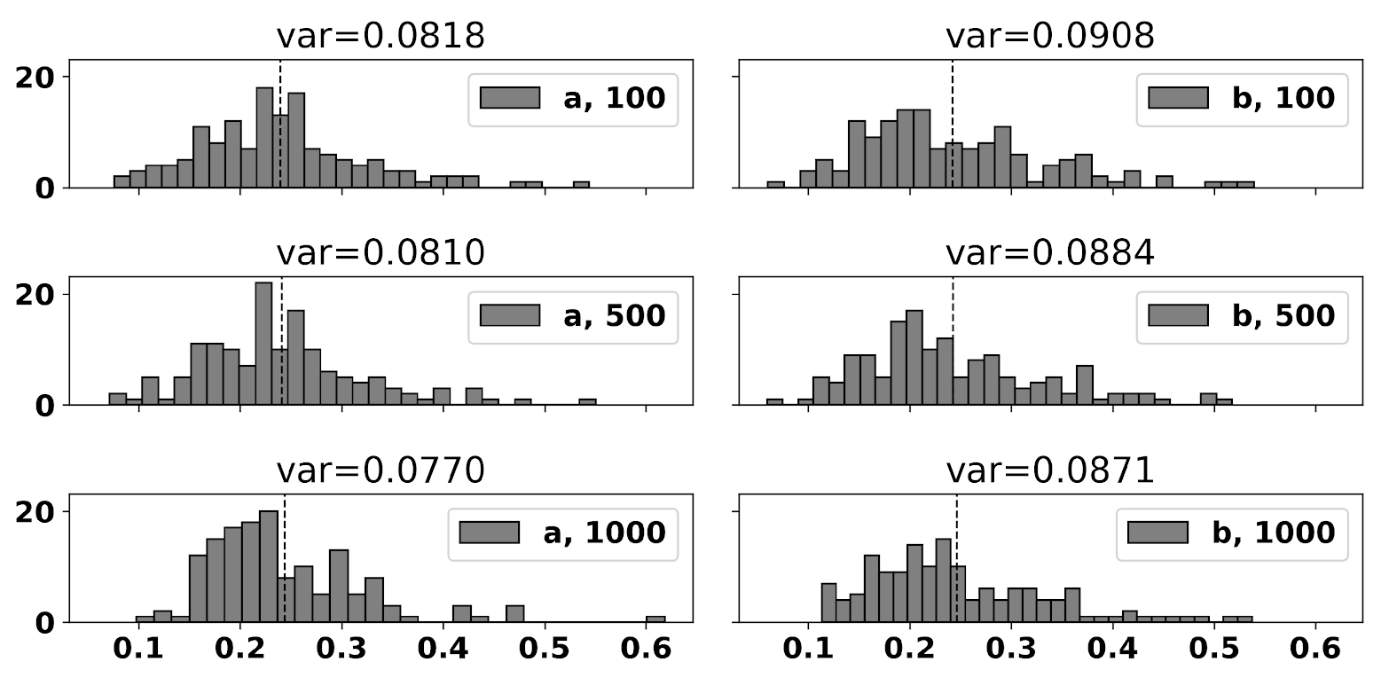}}
    \caption{Predictions of $a$ and $b$ are more stable for larger subgraph samples, indicating a narrowing range effect}
    \label{fig:ab_amazon_node}
\end{figure}

In real-life scenarios, for larger graphs it may take several days to finish even one run of the linear threshold influence maximization. Hence, it is beneficial to use the average of the predicted parameter values generated for the subgraphs on the large-scale original graph. When utilizing subgraph-running results, the more nodes are included in the subgraph samples, the more accurately the average approximates the actual best parameters. Figure \ref{fig:ab_amazon_node} shows this narrowing range effect in response to increase of numbers of nodes in the subgraph samples for the Amazon co-purchasing network.

\begin{figure}[htpb]
    \centering

    \includegraphics[width=\textwidth]{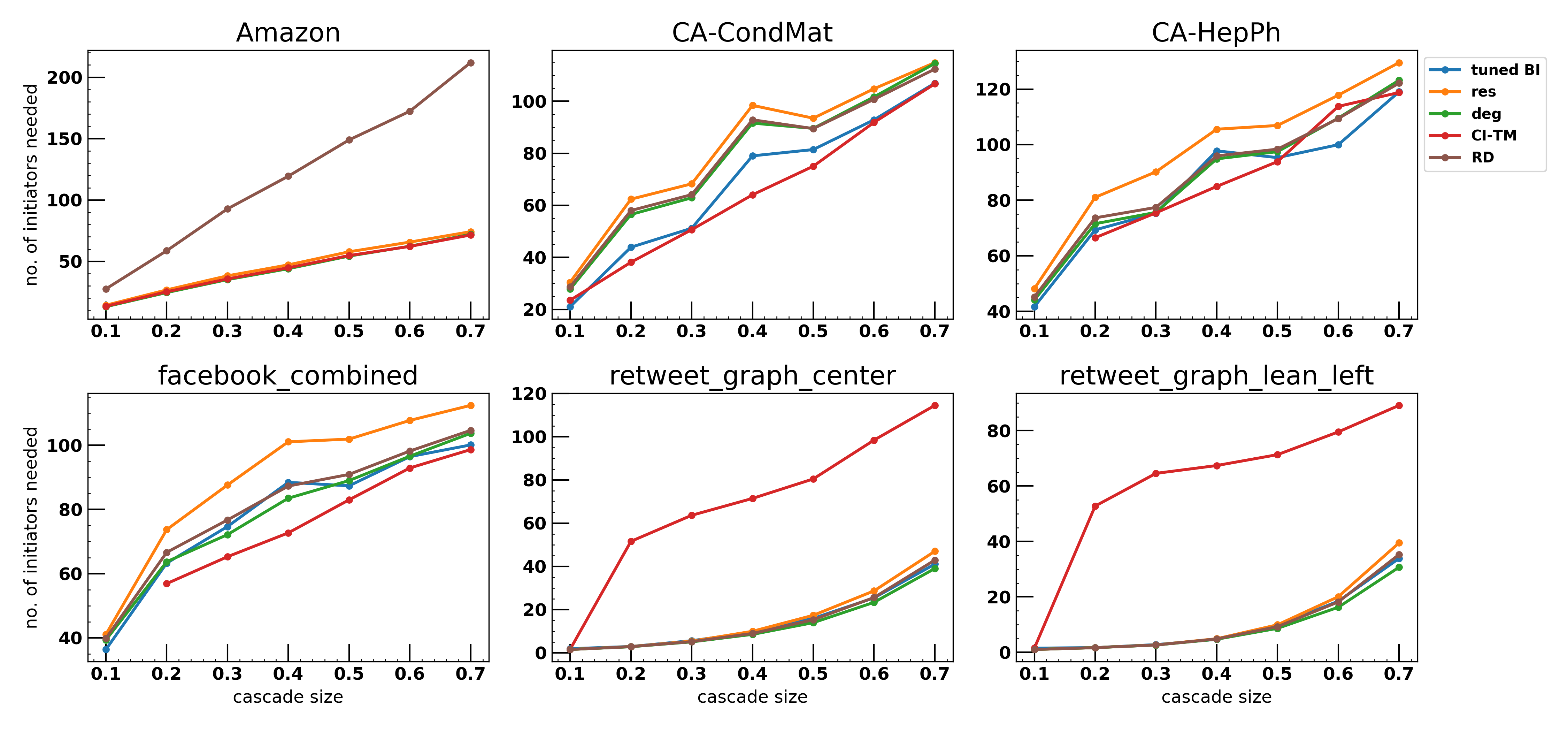}   
    \caption{Comparison of performance of tested heuristics on other real-world net-work subgraphs}

    
    \label{other_network}
\end{figure}

Here, we use other real-world networks for testing, The results are averaged over 1000 network subgraph samples and 50 for others, are summarized in Fig. \ref{other_network} without addressing to specific experiment settings (i.e. the distribution of resistance thresholds) since the individual results were similar to each other. 

For the six real-world networks, the tuned BI approach performs consistently well for all kinds of real-world networks included (i.e. academic collaboration networks and social networks). However, the CI-TM with $L=1$ shows bifurcation behaviours for the Twitter retweet graphs and the others, indicating that neglecting the resistance aspect of a influence propagation system could be detrimental to the performance.


\section{Conclusion}\label{conclusion}
We use synthetic network data to train the Random Forest Classification to tune the BI algorithm parameters for the high performance on influence maximization problem. Our contributions include the following. We identified the most important features for all the BI parameters, of which the threshold $\phi$ distribution standard deviation dominates others. We designed a novel tuned BI heuristic and compared it  with other node indexing heuristics on six real-world networks. The results demonstrate that the tuned BI approach outperforms the other tested heuristics, and reduces the number of needed initiators by up to 10\%. 

\section{Acknowledgement}
This work work was supported in part by the Army Research Laboratory (ARL) through the Cooperative Agreement (NS CTA) Number W911NF-09-2-0053, the Office of Naval Research (ONR) under Grant N00014-15-1-2640, and by the Army Research Office (ARO) under Grant W911NF-16-1-0524. The views and conclusions contained in this document are those of the authors and should not be interpreted as representing the official policies either expressed or implied of the Army Research Laboratory or the U.S. Government.
%
%
\bibliographystyle{unsrt}
\bibliography{mybib.bib}

\end{document}